\pdfoutput=1

\documentclass[11pt]{article}

\usepackage{ACL2023}

\usepackage{times}
\usepackage{latexsym}

\usepackage[T1]{fontenc}

\usepackage[utf8]{inputenc}

\usepackage{microtype}
\usepackage{caption}
\usepackage{subcaption}

\usepackage{inconsolata}

\usepackage{multirow} 
\usepackage{multicol}
\usepackage{graphicx}
\usepackage{mathabx}
\usepackage{enumitem}
\usepackage{xcolor}

\setlist[enumerate,1]{leftmargin=*} 
\setlist[enumerate,0]{label=(\alph*),widest=a}

\usepackage{todonotes}

\usepackage{booktabs,arydshln}

\makeatletter
\def\adl@drawiv#1#2#3{%
        \hskip.5\tabcolsep
        \xleaders#3{#2.5\@tempdimb #1{1}#2.5\@tempdimb}%
                #2\z@ plus1fil minus1fil\relax
        \hskip.5\tabcolsep}
\newcommand{\cdashlinelr}[1]{%
  \noalign{\vskip\aboverulesep
           \global\let\@dashdrawstore\adl@draw
           \global\let\adl@draw\adl@drawiv}
  \cdashline{#1}
  \noalign{\global\let\adl@draw\@dashdrawstore
           \vskip\belowrulesep}}
\makeatother

\title{A Dataset for Physical and Abstract Plausibility\\and Sources of Human Disagreement}

\author{Annerose Eichel, Sabine Schulte im Walde \\
         Institute for Natural Language Processing, University of Stuttgart \\ 
         \texttt{\{annerose.eichel,schulte\}@ims.uni-stuttgart.de}}

\begin{document}

\maketitle

\begin{abstract}
We present a novel dataset for physical and abstract plausibility of events in English. Based on naturally occurring sentences extracted from Wikipedia, we infiltrate degrees of abstractness, and automatically generate perturbed pseudo-implausible events. We annotate a filtered and balanced subset for plausibility using crowd-sourcing, and perform extensive cleansing to ensure annotation quality. In-depth quantitative analyses indicate that annotators favor plausibility over implausibility and disagree more on implausible events. Furthermore, our plausibility dataset is the first to capture abstractness in events to the same extent as concreteness, and
we find that event abstractness has an impact on plausibility ratings: 
more concrete event participants trigger a perception of implausibility. 
\end{abstract}

\section{Introduction}
\begin{figure}[!htbp]
    \centering
    \includegraphics[width=0.5\textwidth]{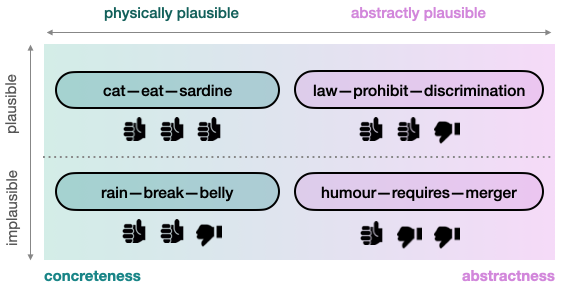}
    \caption{Plausible and implausible example events integrating degrees of concreteness/abstractness, cf. physical (green) and abstract (pink) levels. Annotators might agree (thumbs up) or disagree (thumbs down) on the (im)plausibility of the events. 
    }
    \label{fig:example}
    \vspace{-0.3cm}
\end{figure}

The ability to discern plausible from implausible events is a crucial building block for natural language processing (NLP).
Most previous work on modelling plausibility however focuses on the kinds of semantic knowledge necessary for distinguishing a \textit{physically} plausible event from an implausible one \cite{wang-etal-2018-modeling,porada-etal-2019-gorilla}. As illustrated in Fig.~\ref{fig:example}, the current study extends the traditional focus to discern physically plausible events such as \textit{cat-eat-sardine} from physically implausible ones such as \textit{rain-break-belly}. Furthermore, while recent datasets include some events with conceptually \textit{abstract} participants \citep{emami-etal-2021-adept,pyatkin-etal-2021-possible},
as to our knowledge no dataset nor model up to date has paid attention to the interaction of event plausibility and abstractness of the involved concepts. 
We propose to systematically examine plausibility across levels of abstractness, and distinguish between abstractly plausible events such as \textit{law-prohibit-discrimination} and abstractly implausible ones such as \textit{humour-require-merger}. We hypothesize that (i) plausible vs. implausible events can be captured through physical vs. abstract levels, and that (ii) integrating degrees of abstractness into events fosters the understanding and modelling of plausibility
(cf. Fig.~\ref{fig:example}).

We start out with a set of attested, i.e., \textit{plausible}, natural language events in form of \textit{s-v-o} triples from the English Wikipedia, assign abstractness ratings to event participants, and partition the triples into bins with varying degrees of abstractness. We then automatically generate pseudo-\textit{implausible} event triples and assign degrees of abstractness in a similar way. To obtain human plausibility ratings for each event triple, we conduct a crowd-sourcing annotation study. We collect and validate a total of 15,571 judgements amounting to an average of 8.9 ratings for 1,733 event triples.

Human intuition regarding the assessment of plausibility is, however, incredibly multi-faceted, highly individual, and not easily reproducible automatically \cite{resnik1993}. 
In particular, boundaries between categories to be annotated or predicted might not necessarily be strictly true \textit{or} false, i.e., either plausible or implausible, thus reflecting the true underlying distribution of non-deterministic human judgements with inherent disagreement about labels \cite{baan-etal-2022-stop}. Over the past decade, a growing body of work has emphasized the need to incorporate such disagreement in NLP datasets to reflect a more realistic and holistic picture across NLP tasks \cite{plank-etal-2014-linguistically,aroyo2015truth,jamison-gurevych-2015-noise,basile-etal-2021-need,uma-etal-2021-semeval}. Accordingly, we argue for the necessity to preserve and examine disagreement when annotating and modelling plausibility, and represent inherent disagreement in annotation in order to devise a range of silver standards for analysis and modelling. More specifically, we disentangle subjectivity from annotation error, limitations of the annotation scheme, and interface \cite{pradhan-etal-2012-conll,poesio-etal-2019-crowdsourced}, and examine disagreements in physical and abstract plausibility annotation. 

Overall, we find that our annotators tend to favor plausibility over implausibility, and we observe stronger disagreements for implausible in comparison to plausible events. Second, we explore the impact of abstractness on plausibility ratings. Here, our results reveal a positive relation between plausibility and events consisting of more abstract words, while implausibility is mostly found in predominantly concrete events.

\section{Background and Related Work} 

\subsection{Capturing (Semantic) Plausibility}
The notion of plausibility has been approached from many perspectives. Inspired by the overview in \citet{porada-etal-2021-modeling}, we present distinctions and discuss viewpoints from previous work.
Similarly to related notions such as \textit{selectional preference} \cite{wilks1975,resnik1993,erk-etal-2010-flexible,van-de-cruys-2014-neural,zhang-etal-2019-sp,metheniti-etal-2020-relevant} and \textit{thematic fit} \cite{chersoni-etal-2016-towards,sayeed-etal-2016-thematic,pedinotti-etal-2021-cat}, plausibility estimations capture non-surprisal in a given context. For example, the event \textit{kid-sleep} with the agent \textit{kid} is less surprising than \textit{tree-sleep} and therefore considered more plausible. Within the context of (semantic) plausibility, however, plausible events are not necessarily assumed to be the most typical or preferred events. This stands in contrast with selectional preference or thematic fit, where whatever is not preferred is considered atypical albeit, in principle, a given event might be plausible. \citet{wilks1975} also discusses naturally occurring cases where the most preferred option does not yield the only correct interpretation: ``[t]he point is to prefer the normal, but to \textit{accept} the unusual.''

In this vein, \citet{wang-etal-2018-modeling} propose the task of semantic plausibility as ``recognizing plausible but possibly novel events'', where a ``novel'' event might be an unusual but nevertheless plausible event. \citet{porada-etal-2021-modeling} further point out that ``[p]lausibility is dictated by likelihood of occurrence on the world rather than text'', and attribute this discrepancy to reporting bias \cite{gordon-vandurme-2013, shwartz-choi-2020-neural}. For example, it is much more likely that the event \textit{human-dying} is attested than the event of \textit{human-breathing}. The sum of all plausible events in a given world thus encompasses not only the sum of all attested events in a corpus (including modalities other than text), but also possibly plausible events which are \textit{not} necessarily attested in a corpus. 

In our definition what is \textit{preferred} is considered the \textit{most plausible}, but what is \textit{unusual} might still be highly \textit{plausible}. Plausibility therefore (i) \textbf{exceeds the boundaries of (selectional) preference}. Further, plausibility 
(ii) is \textbf{a matter of degree} as the preferred is considered \textit{more} plausible. In turn, what is unusual is still considered plausible albeit to a lesser degree.
Moreover, plausibility (iii) captures \textbf{non-surprisal in a given context}, and (iv) denotes what is generally \textbf{likely, but not necessarily attested in a given corpus}. 


\subsection{Measuring Semantic Plausibility}

There are various positions on how to model, measure, and evaluate whether an event triple is plausible.
In this study, we model plausibility as the proportion of what is considered plausible, requiring a minimal label set of \{implausible, plausible\} \cite{wang-etal-2018-modeling}. Note that a value regarding what is ``true'' is not involved in measuring plausibility. Consider the examples \textit{eat-strawberry}, \textit{eat-pebble}, and \textit{eat-skyscraper}. Given our label set, the first two events would be considered plausible (even though they strongly vary in their degree of plausibility and likelihood to be attested in text with eating a strawberry considered more plausible than the less, but still plausible process of eating a pebble)\footnote{Using e.g., Google n-grams, \textit{eat-strawberry} is clearly attested more often than \textit{eat-pebble}, while \textit{eat-skyscraper}
is not attested at all.}, while the last event is physically implausible. Derived label sets such as \{implausible, neutral, plausible\} may include a ``neutral'' label which is considered to not carry plausibility information, as it does not provide insight into whether an expression is (im)plausible \cite{anthonio-etal-2022-clarifying}.

When annotating plausibility, drawing hard lines between labels is difficult and increases in complexity when considering words and concepts that are more abstract than concrete.
This is especially true when considering free-standing events where no information on limiting factors regarding interpretation can be inferred. An example would be \textit{human-breathe} which is plausible 
unless the human in question is dead. A more complex example would be \textit{human-have-human\_rights}, which is likely to be considered plausible 
by the majority of people and mirrored by corresponding laws in many countries, but (a) not universally accepted by each individual, and (b) not formalized as such by all countries. 


\subsection{Physical and Abstract Plausibility}

Concepts can be described in accordance with the way people perceive them. While concepts that can be seen, heard, touched, smelled, or tasted are described as \textit{concrete}, those that cannot be perceived with the five senses are referred to as \textit{abstract} \cite{barsalou2005situating,brysbaert2014}. Examples of concrete concepts include \textit{apple, house} and \textit{trampoline}, abstract examples encompass \textit{absurdity, luck}, and \textit{realism}. While instances at each extreme of abstractness occur, the notion is not binary but rather continuous, including many concepts between each extreme. Mid-range examples include concepts such as \textit{inflation, punctuality} and \textit{espionage}.

The grounding theory of cognition argues that humans process abstract concepts by creating a perceptual representation that is inherently concrete as it is generated through exposure to real world situations using our five senses \citep{vandam2010,brysbaert2014}.
However, more recent work brings forth evidence suggesting that such representations incorporate both perceptual and non-perceptual features \citep{dove2009,naumann-etal-2018-quantitative,frassinelli-schulte-im-walde-2019-distributional}.

Regarding suitable abstractness ratings, we find a variety of datasets of growing size and diversity for many languages.\footnote{See for a detailed overview, e.g., 
\citet{SchulteImWalde/Frassinelli:22, charbonnier-wartena-2019-predicting}.} A widely used collection are the concreteness norms devised by \citet{brysbaert2014}, who collected ratings for approx. 40K ``generally known'' English words such as \textit{sled} and \textit{dream}, referring to strength of sense perception.

\subsection{Disagreement in Dataset Construction} 

While humans excel at assessing plausibility, they might naturally disagree regarding the plausibility of an event such as \textit{law-prohibit-discrimination}. In the course of the last decade, a growing line of research argues for the preservation and integration of disagreement in dataset construction, modelling, and evaluation 
\cite{aroyo2015truth,pavlick-kwiatkowski-2019-inherent,basile-etal-2021-need,fornaciari-etal-2021-beyond,uma-etal-2021-semeval}\footnote{For an overview we refer to, e.g., \citet{basile2021,Uma2021LearningFD}.}. While highly subjective tasks such as sentiment analysis \cite{yin-etal-2012-unifying,kenyon-dean-etal-2018-sentiment} and offensive language detection \cite{leonardelli-etal-2021-agreeing,almanea-poesio-2022-armis} have gathered particular attention, prior work has also presented evidence for seemingly objective tasks requiring linguistic knowledge such as PoS tagging \cite{gimpel-etal-2011-part,hovy-etal-2014-experiments,plank-etal-2014-linguistically}. 
We thus argue for the necessity to disentangle, devise, and examine disagreement when annotating and modelling plausibility. In contrast to previous work on plausibility assessments, we represent inherent disagreement in annotation and devise a range of silver standards for analysis and modelling.

\begin{figure*}[!htbp]
    \centering
    \includegraphics[width=\textwidth]{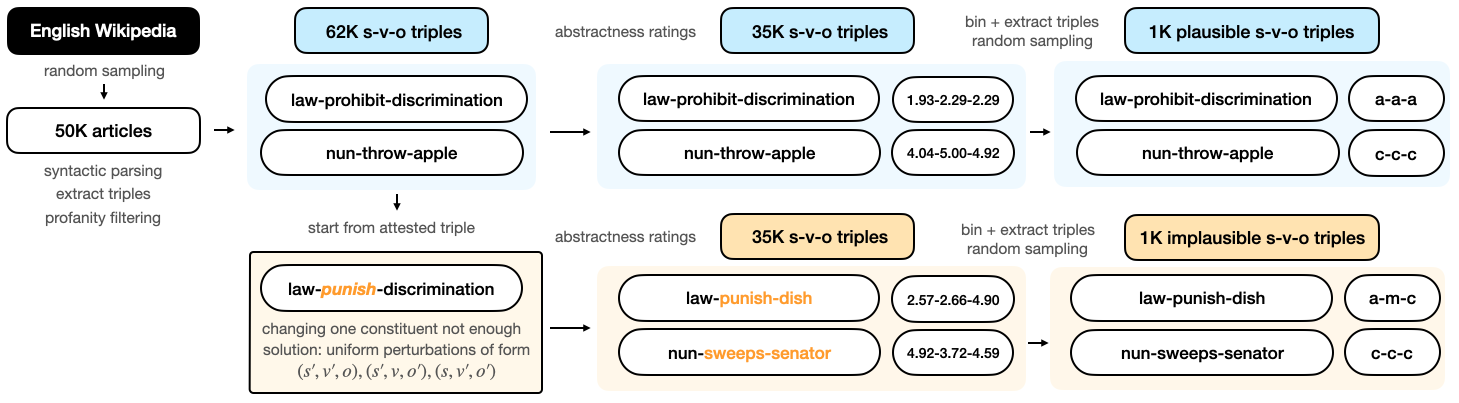}
    \caption{Simplified illustration of dataset construction starting with the extraction of attested event triples from a sample of the English Wikipedia. We filter triples, assign abstractness ratings, bin, and sample 1,080 plausible event triples for 27 abstractness combinations (marked in blue). Based on attested triples, we automatically generate pseudo-implausible triples and similarly filter triples, assign abstractness ratings, perform bining, and sample 1,080 implausible event triples (marked in yellow).
    }
    \label{fig:dataset}
    \vspace{-0.3cm}
\end{figure*}

\section{Construction of Event Targets} \label{sec:data}

Our first goal is to create a dataset\footnote{The dataset including event triples, ratings, and aggregated labels is available at \url{https://github.com/AnneroseEichel/PAP}.} that systematically (a) covers both plausible event triples that are selectionally preferred or unusual, (b) captures events attested in the real world, i.e., extracted from triples produced in natural language, (c) measures plausibility on a degree scale from plausible to implausible, and (d) puts equal emphasis on both abstractly and physically plausible events. We visualize the dataset construction process in Fig.~\ref{fig:dataset}. 

\subsection{Extracting Natural Language Triples}

To compile a set of natural language triples, we first extract all text from an English Wikipedia dump using \texttt{gensim} \cite{rehurek2010}. 
We then randomly sample $k$ articles\footnote{We only store the section texts and discard (section) titles as they tend to not contain 3-tuples of the desired form.} with $k$=50,000 and syntactically parse the articles using \texttt{stanza} \cite{qi2020stanza}. 
Next, we extract a triple $(s,v,o)$ whenever the following conditions are satisfied: $s$ is the lemma of the head of \texttt{nsubj}, $o$ is the lemma of the head of \texttt{obj}, and $v$ is the lemma of the head of the root verb. 
We only allow nouns in subject and object positions and disregard proper names and pronouns as well as nouns and verbs that are
part of a compound, yielding 62,843 triples. 
We extract each triple once, keeping track of frequency w.r.t sampled text data. Triples containing nouns or verbs that are explicit
or have offensive connotations are filtered out using existing tools.\footnote{filter-profanity, alt-profanity-check} In total, this leaves us with 62,473 triples. 

\subsection{Creating Physically and Abstractly Plausible Triples} \label{subsec:plausi}

To discern triples containing highly concrete words from triples which encompass more abstract words, we assign abstractness scores to all nouns and verbs in a triple,
drawing on the concreteness ratings by \citet{brysbaert2014}. 
We use a reduced collection\footnote{For details on the filtering process, cf. App.~\ref{appsec:dataset_nltriples}.} encompassing 12,880 noun and 2,522 verb targets to assign concreteness ratings to all 62,473 triples where a rating $r$ exists for each word
$w \in \{s,v,o\}$.
Instances with nouns or verbs for which no rating exists are discarded. Overall, the assignment step yields 35,602 triples\footnote{Subject and object types amount to 4,140 and 4,551 unique words, respectively, while verb types are significantly less diverse (1,218 unique words).} with ratings. 
As we are specifically interested in distinctive features of abstractness vs. concreteness and cases which can be found in the middle of the continuous scale, we partition each constituent and each triple into 5 bins [\textit{highly abstract}, \textit{abstract}, \textit{mid-range}, \textit{concrete}, \textit{highly concrete}]. To construct our dataset, we then only consider the bins at each extreme as well as the mid-range bin.  
Each constituent of a triple $t$ can be either \textit{highly abstract} (a), \textit{mid-range} (m), or \textit{highly concrete} (c). Taking the Cartesian product, we thus define 27 possible triple combinations, e.g., triples consisting of words with very high concrete ratings only, e.g., $(c,c,c)$ or fully mixed triples, e.g., $(c,m,a)$. 
To extract triples satisfying the conditions of each of the 27 possible triple combination, we carry out the following steps: 

\begin{enumerate}[leftmargin=*]
    \item Partition each constituent in $s,v,o$ in each $triple_{1...n}$ into 5 bins of equal size,  ranging from very abstract to very concrete.
    Whenever the relative threshold $\theta$ between bins prohibits perfectly equal sizes, we trade perfect bin size for perfectly separated abstractness ratings.
    \item Extract all triples satisfying the conditions of a combination e.g., $(c,c,c)$ from our set of 35,602 triples. 
\end{enumerate}
%
The distribution of all naturally occurring triples for each triple combination 
$\in \{(a,a,a), ... (c,c,c)\}$ is presented in Fig.~\ref{fig:distributions}, App.~\ref{appsec:dataset_distribution}.
To select plausible triples for annotation, we randomly sample 40 triples for each combination, yielding a total of 1,080 plausible triples.

\subsection{Constructing Physically and Abstractly Implausible Triples}
To construct implausible triples, we use the 35,602 cleaned triples for which an abstractness rating as provided by \citet{brysbaert2014} exists.
This restriction makes the task of implausible triple generation non-trivial as the set of possible constituents in each function is now limited to subjects, verbs and objects that are attested to be plausible in their given function. 
Generating perturbations of attested triples as used by \citet{porada-etal-2021-modeling} --where only one constituent, e.g., the subject, is perturbed while verb and object are kept-- also results in disproportionally many plausible triples, e.g., \textit{jurisdiction-evaluate-reaction}.

We thus use only the following perturbations:
For each $t \in$ attested triples, we obtain a randomly perturbed $t^{\prime}$ serving as a pseudo-implausible natural language triple. We uniformly generate perturbations of the form $(s^{\prime}, v^{\prime}, o)$, $(s^{\prime}, v, o^{\prime})$ and $(s, v^{\prime}, o^{\prime})$, where $s^{\prime}$, $v^{\prime}$, and $o^{\prime}$ are arguments randomly sampled from the plausible triple collection taking into account corresponding functions, e.g., only words for which the use as object is attested in the corpus are randomly sampled as an object perturbation.
We discard all triples that exist in the plausible triple collection and only keep unique instances, thus yielding 35,600 pseudo-implausible triples. After profanity filtering, we are left with 35,447 triples. 
We assign abstractness ratings and apply the binning method as described in the previous section~\ref{subsec:plausi}. 

The distribution of physically and abstractly pseudo-implausible triples per combination is shown in Fig.~\ref{fig:distributions} (b), App.~\ref{appsec:dataset_distribution}. In analogy to plausible triple construction, we sample 40 triples for each abstractness combination to obtain 1,080 implausible triples.

\section{Human Annotation} \label{sec:human_annotation}

Our second goal targets the annotation of the collected event triples with respect to subjective assessments of plausibility on a degree scale (1--5) ranging from implausible to plausible. For this, we perform a human annotation study.

\begin{figure*}[!htpb]
     \centering
     \begin{subfigure}[b]{0.43\textwidth}
         \centering
         \includegraphics[width=\textwidth]{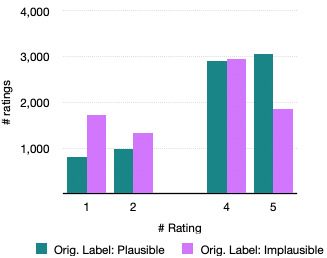}
         \caption{Number of ratings per rating option.}
         \label{fig:distribution_plausible_ts}
     \end{subfigure}
     \hfill
     \begin{subfigure}[b]{0.43\textwidth}
         \centering
         \includegraphics[width=\textwidth]{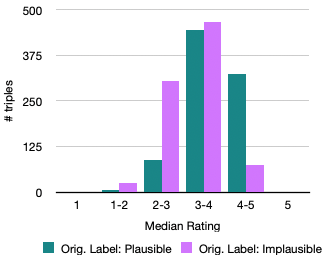}
         \caption{Number of triples per average median rating bin.}
         \label{fig:distribution_implausible_ts}
     \end{subfigure}
     \hfill
        \caption{(a) Number of plausibility ratings per rating option where ratings below 3 denote implausibility and ratings above 3 denote plausibility. (b) Number of triples across ratings aggregated as averaged median ratings. Ratings range from implausible $\{1,2\}$  to plausible $\{4,5\}$.
        }
        \label{fig:rating_insights}
\end{figure*}

\subsection{Collecting Ratings for (Im)Plausibility}

\paragraph{Task} We collect plausibility judgements on Amazon Mechanical Turk\footnote{\url{https://www.mturk.com/}}
for our 2,160 plausible and implausible triples. Each triple is annotated by 10 annotators. In particular, we ask annotators to indicate whether a given sentence is implausible or plausible using a sliding bar (corresponding to a scale from 1 to 5). An example of the task with full instructions as presented to annotators in our Human Intelligence Task (HIT) is illustrated in Fig~\ref{fig:hit_example}, App.~\ref{appsec:human_annotation_interface}. To avoid bias, the slider is by default set to the middle of the bar. Annotators are required to move the slider and thereby make a decision for either plausible or implausible. Task instructions clearly inform about the possibility of submission rejections if the slider remains in the middle position. 



\paragraph{Annotators} 
Participation is limited to annotators based in the United States and the United Kingdom.
We further require annotators to have a HIT Approval Rate $>$ 98\% and a number of $\geq$ 1,000 approved HITs from previous work.

\paragraph{Quality Checks} 
To track annotation quality, we use an initial set of 20 manually produced check instances (cf. App.~\ref{appsec:human_annotation_check}) that were judged clearly plausible/implausible by the authors and an additional English native speaker. Annotators are presented batches of 24 randomly shuffled plausible or implausible triples, plus one randomly sampled check instance. In case of failed check instances, we discard all annotations submitted by the corresponding worker.

\subsection{Annotation Post-Processing}
After discarding submissions where the slider is set to the default (rating$=$3) as well as submission from workers who failed a check instance, we collect a total of 21,317 plausibility ratings.\footnote{After our main collection round, we experiment with basic post-processing steps and perform a small second collection round to collect 5 ratings for 84 triples with $<$8 ratings.} 
We further perform the following post-processing steps in order to minimise the impact of spam and low-quality annotations
regarding the plausibility of a given event \cite{roller-etal-2013-un,rodrigues2017,leonardelli-etal-2021-agreeing}, with datasets statistics at every processing step shown in Table~\ref{tab:annotation_stats}, App.~\ref{appsec:human_annotation_post}. We first filter out ratings from workers who submitted annotations for $<$10 instances. Assuming that events observed in Wikipedia represent plausible events, we then exclude ratings from workers whose annotations disagree with the original label \textit{plausible} in more than 75\% of their corresponding submissions. 

After these steps, our number $n$ of annotators $A$ still amounts to a large set of $n$A $>$ 500 annotators. 
To ensure sufficient agreement between annotators, we calculate a \textit{soft} pairwise Jaccard Coefficient $J$ \cite{jaccard1902}\footnote{See App.~\ref{appsec:human_annotation_post} for details on the calculation.} for all annotator combinations, and only keep annotations from workers whose submissions yield an average $J>0.4$, following \citet{bettinger-etal-2020-domain}. Finally, 
we keep only triples in the dataset if they received at least 8 ratings.

\subsection{Dataset Statistics}
After post-processing, we are left with 15,571 plausibility ratings
for 1,733 triples (80\% of the original triple set). With respect to instance coverage per abstractness combination, we have an average number of 32 triples per combination for both plausible and implausible triples with a minimum of 27 triples for the combinations $(a,a,m)$ and $(m,c,c)$ for plausible and implausible triples, respectively. Triples receive between 8 and 12 ratings, with an average of 8.9 ratings.

Estimated average Inter-Annotator Agreement (IAA) across our post-processed dataset using the previously introduced soft pairwise Jaccard Coefficient reaches 0.64. This indicates reasonable agreement among annotators; cases of disagreement we will explore in the next section.


\section{Analysis of Human Judgements and Disagreement} \label{sec:human_analysis}

\subsection{Examining Rating Distributions}
Fig.~\ref{fig:rating_insights} (a) shows the distribution of ratings across the four rating options, with green and pink bars indicating originally plausible and implausible label, respectively. The distribution is skewed towards plausibility with 68.98\% ratings $\in \{4,5\}$. We aggregate all individual ratings as average median rating per triple and show the resulting distribution in Fig.~\ref{fig:rating_insights} (b). While the distribution for originally plausible triples (green bars) evens out as expected with a peak number of average median rating for average plausibility (avg. median ratings $\in (3;4]$), a similar peak can be observed given the distribution for originally implausibly triples (pink bars). The graph also shows differences, namely substantially more triples with a median rating indicating weak implausibility (avg. median rating $\in (2;3]$) for originally implausible triples. On the other hand, high plausibility (rating $\in (4;5]$) is annotated for mostly originally plausible triples. 

\begin{figure*}[!htpb]
     \centering
     \begin{subfigure}[b]{0.48\textwidth}
         \centering
         \includegraphics[width=\textwidth]{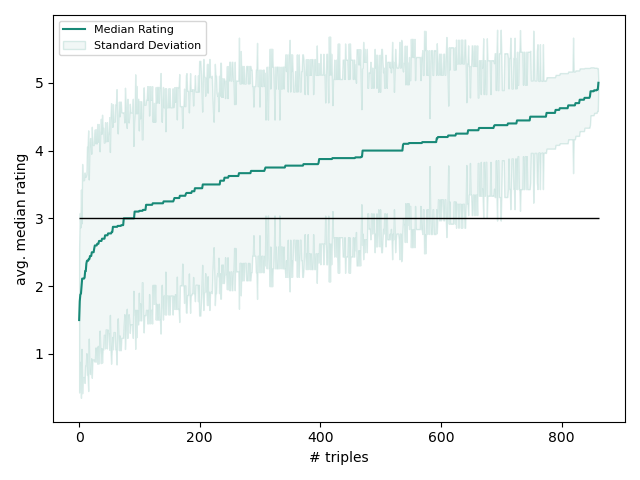}
         \caption{Average median rating across plausible triples.}
         \label{fig:plausible_sdline}
     \end{subfigure}
     \hfill
     \begin{subfigure}[b]{0.48\textwidth}
         \centering
         \includegraphics[width=\textwidth]{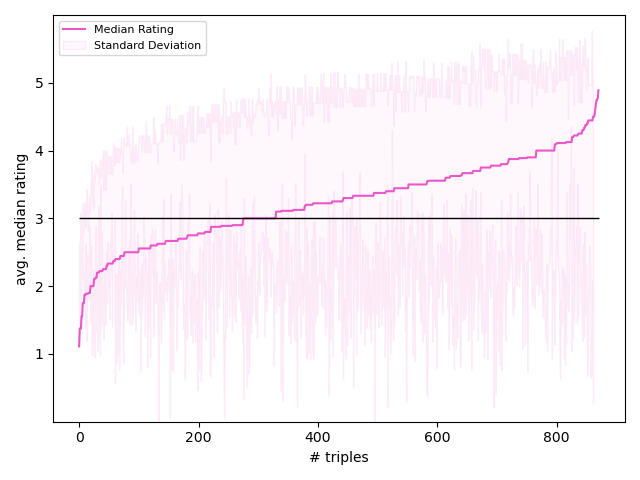}
         \caption{Average median rating across implausible triples.}
         \label{fig:implausible_sdline}
     \end{subfigure}
     \hfill
        \caption{
        Average median ratings across originally plausible (a) and implausible (b) triples with standard deviation visualized as cloud around average rating lines.
        Triples are represented numerically on the x-axis. The black horizontal line denotes a median rating of 3. Average median ratings for \textit{plausible} triples \textit{below} the line disagree with the original label, while the opposite is true for average median ratings for \textit{implausible} triples. Here, ratings \textit{above} the line disagree with the original label.}
        \label{fig:mean_rating}
\vspace{-0mm}
\end{figure*}

To further investigate the skew towards plausibility, we visualize the average median rating for originally plausible and implausible triples in Fig.~\ref{fig:mean_rating}. The plot also illustrates the standard deviation of the values as a cloud. We observe that annotator ratings tend to show more overlap for plausible triples, with standard deviation decreasing with higher plausibility. In contrast, rating triples labeled as implausible result in greater deviation from the average mean rating decreasing only with implausibility. Taking into account the black horizontal line at a median rating of 3, we clearly see that median ratings for originally plausible triples are mostly above the cut line, thus indicating an overlap with the original label. On the other hand, median ratings for originally implausible triples are mostly below the cut line, thus indicating a clash with the original label.

These observations suggest (i) that \textbf{humans favor plausibility over implausibility}, while avoiding the extreme on the plausibility end of the scale, and (ii) that \textbf{implausibility yields higher disagreement}, as annotators disagree more when rating triples that were originally labeled as implausible. 

\subsection{Exploring the Impact of Abstractness on Plausibility Ratings}

\begin{figure*}[!htpb]
     \centering

     \begin{subfigure}[b]{0.48\textwidth}
         \centering
         \includegraphics[width=\textwidth]{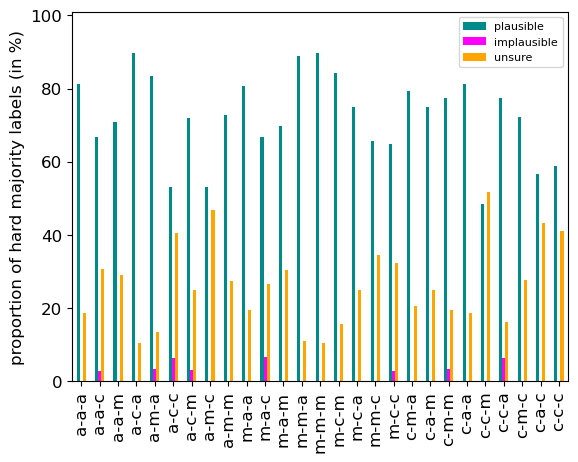}
         \caption{Attested plausible triples.}
         \label{fig:distribution_plausible_ts}
     \end{subfigure}
     \hfill
     \begin{subfigure}[b]{0.48\textwidth}
         \centering
         \includegraphics[width=\textwidth]{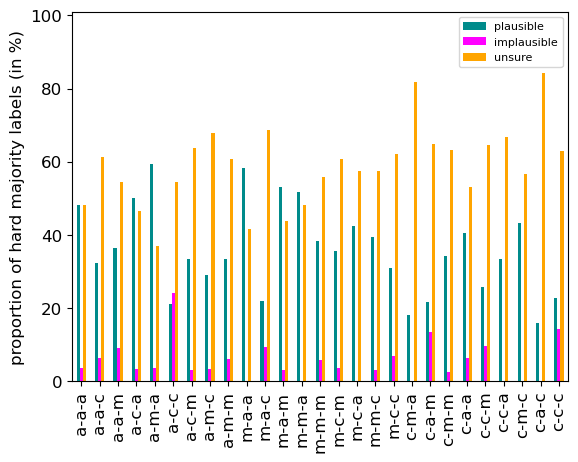}
         \caption{Perturbed implausible triples.}
         \label{fig:distribution_implausible_ts}
     \end{subfigure}
     \hfill
        \caption{Proportion of
        strict majority ratings ($\geq$70\%) across abstractness combinations for attested plausible triples (a) and perturbed implausible triples (b). Green bars denote a majority of plausible ratings $\in \{4,5\}$, pink bars refer to a majority of implausible ratings $\in \{1,2\}$, and orange bars capture cases of no clear majority.}
        \label{fig:majority}
\vspace{-2mm}
\end{figure*}

\paragraph{Abstractness at Event Level} \label{para:event_level}To assess the relation between degrees of abstractness for combinations of words and plausibility on physical and abstract levels, we first examine the proportion of plausibility ratings across triples from each of our 27 abstractness combinations. For this, we calculate a 
\textit{strict} majority ($\geq$70\%) for each triple. Whenever ratings do not point to a majority, i.e., \textit{50\% plausible} vs. \textit{50\% implausible}, we mark the triple as \textit{unsure}. We present a visualization in Fig.~\ref{fig:majority} where green bars denote a strict majority of plausible ratings $\in \{4,5\}$, pink bars refer to a strict majority of implausible ratings $\in \{1,2\}$, and orange bars illustrate the lack of clear majorities.

For attested plausible triples, original label and proportional majority rating overlap in all cases. In only three cases we observe majority ratings proportions below 50\%, namely for the mostly concrete combinations $(c,c,m)$, $(a,c,c)$, and $(a,c,m)$. In contrast, majority rating proportions are generally higher for more abstract combinations, e.g., $(a,a,a)$, $(m,a,a)$. While a very low average of majority ratings for implausibility (1.3) can be observed, an average of 26.2 is obtained for triples with no majority. These observations suggest that (i) implausibility is most likely assigned to triples with concrete words, inducing higher disagreement among annotators, (ii) plausibility is most likely assigned given more abstract words.

For perturbed implausible triples, the picture looks different with only one abstractness combination for which original and majority rating proportions overlap, namely $(a,c,c)$. 
For four highly abstract combinations $(a,m,a)$, $(m,a,a)$, $(m,m,a)$, $(m,m,a)$, a plausible majority is observed. However, in comparison with attested plausible triples, disagreement and uncertainty is much higher with no clear majority for 80\% of abstractness combinations.
These findings underline the observations for attested plausible triples with (i) implausibility being easier to catch given concrete words, and (ii) plausibility connected to more abstract words.


\paragraph{Abstractness at Event Constituent Level} We further examine abstractness at constituent level, i.e., we explore whether abstractness degrees of individual constituents play a role. For this, we again calculate strict majority ratings across triples for each abstractness combination in a binary label setup (cf. \ref{para:event_level}). We focus on triples with a $\geq$ 70\% majority for either plausible or implausible and calculate the proportion of concrete, mid-range, and abstract constituents $\in \{s,v,o\} \in t$.

Results are presented in Table~\ref{tab:constituent_analaysis}.
For constituents of triples receiving plausible majority votings, no particular pattern stands out: we find relatively equal shares for all constituents across abstractness levels. For originally implausible triples rated plausible, we observe a slightly higher share of mid-range and abstract constituents. In contrast, abstractness levels seem to play a more important role for constituents of triples with implausible majority votings. For both originally plausible and implausible triples, percentage shares clearly increase for concrete subjects and objects as compared to triples with plausible majorities. We also observe more abstract verbs, while shares of concrete and mid-range verbs decrease. In addition, a decrease in abstract subjects and objects as well as mid-range subjects can be observed. Regarding verb constituents, the line seems to be clear-cut between verbs as we find an increase in abstract, a decrease in mid-range, and relatively equal shares for concrete verbs.

These examinations suggest that abstractness levels of event constituents are especially important when assessing the absence of plausibility. Generally, events with a majority voting for implausible tend to include more concrete subjects and objects. However, the picture gets more diverse with clear increases in abstract verbs. Interestingly, these observations hold irrespective of the original label. 

The exploration of abstractness at event constituents underlines our findings from the previous analysis focusing on abstractness at event level. We again find that the majority of human annotators tend to agree on what is plausible, while implausibility seems to be harder to catch and introduces more disagreement. Moreover, assignment \textbf{likelihood of plausibility increases with abstractness} of triple constituents, whereas assignment \textbf{likelihood of implausibility increases with concreteness} of triple constituents -- no matter the underlying original label.

\section{Final Dataset: Aggregations}

To foster learning with and from disagreement, we release not only (i) the raw annotator ratings, but also (ii) provide the following standard aggregations to enable various perspectives for interpretation and modelling; for further aggregation options see e.g., \citet{Uma2021LearningFD}. We account for both multi-class (label $\in \{1,2,4,5\}$) and binary (label either plausible $\in \{4,5\}$ or implausible $\in  \{1,2\}$) categorizations. The dataset is available at \url{https://github.com/AnneroseEichel/PAP}.

\begin{table*}[]
\small
\centering
\begin{tabular}{@{}lrrrrrr@{}}
\toprule
among 1,733 valid triples & \multicolumn{3}{l}{originally plausible}                & \multicolumn{3}{l}{originally implausible}              \\ \cmidrule(l){2-7} 
 & maj. plausible & maj. implausible & no maj. & maj. plausible & maj. implausible & no maj. \\ \midrule
                 \# triples & 622      & 11       & 229 & 309     & 46      & 516 \\ \midrule \midrule 
                 constituent            & \multicolumn{6}{c}{constituent proportion (in \%)} \\ \midrule 
concrete subjects & 0.106        & \textbf{0.182}       & \textbf{0.158}       & 0.100        & \textbf{0.152}        & 0.144        \\
concrete verbs     & 0.109        & 0.091       & \textbf{0.162}       & 0.093        & 0.116        & \textbf{0.172}        \\
concrete objects   & 0.100        & \textbf{0.182}       & 0.070        & 0.088        & \textbf{0.159}        & 0.060        \\
mid-range subjects & \textbf{0.115}        & 0.061       & 0.129       & 0.118        & 0.058        & 0.144        \\
mid-range verbs    & \textbf{0.115}        & 0.091       & 0.034       & \textbf{0.128}        & 0.072        & 0.132        \\
mid-range objects  & 0.111        & 0.061       & 0.125       & 0.111        & 0.138        & 0.042        \\
abstract subjects  & 0.113        & 0.091       & 0.148       & 0.115        & 0.123        & \textbf{0.146}        \\
abstract verbs     & 0.109        & 0.152       & 0.148       & 0.112        & 0.145        & 0.129        \\
abstract objects   & \textbf{0.122}        & 0.091       & 0.025       & \textbf{0.134}        & 0.036        & 0.031        \\ \bottomrule
\end{tabular}
\caption{Overview of constituent analysis focusing on triples with a $\geq$ 70\% majority (maj.) for either plausible or implausible triples (\# triples). We present the proportion of concrete, mid-range, and abstract constituents $\in \{s,v,o\} \in t$ for each abstractness level (concrete, mid-range, abstract) and constituent (subject, verb, object), in~\%. For completeness, we also show constituent proportions for triples with no strict majority (no maj.).}
\label{tab:constituent_analaysis}
\vspace{-3mm}
\end{table*}

\begin{enumerate}
    \item \textbf{Strict Majority with Disagreement}\\
    Classes are assigned based on a 70\% majority for a multi-class or binary setup. In case of no clear majority, a label denoting disagreement is assigned to reflect conflicting perspectives of annotators.
    \item \textbf{Distribution}\\
    To account for fine-grained disagreement and uncertainty, we calculate class distributions for a multi-class or binary setup.
    \item \textbf{Probabilistic Aggregation}\\
    As we work with crowd workers, we also provide probabilistic label aggregations using Multi-Annotator Competence Estimation (MACE)\footnote{\url{https://github.com/dirkhovy/MACE}} \citep{hovy-etal-2013-learning}. MACE leverages an unsupervised item-response model that learns to identify trustworthy crowd annotators and predicts the correct underlying label. We provide both predicted silver labels and class distributions for a multi-class and binary setup.
\end{enumerate}

\section{Discussion} \label{sec:results}

We formulated the task of automatically distinguishing \textit{abstract} plausible events from implausible ones as an extension of \citet{wang-etal-2018-modeling} who focused specifically on \textit{physical} plausible events. Based on the presented findings, we affirm our hypothesis as to (i) whether plausible and implausible events can be systematically captured on physical and abstract levels by (ii) integrating degrees of abstractness for combinations of words. 

We further note differences in collected annotations with assignment likelihood of plausible ratings increasing with abstractness of events' constituents, while concreteness seems to facilitate the detection of more implausible events. We hypothesize that more concrete words evoke a more stable mental image grounded in the real world. Events like our introductory example \textit{rain-breaks-belly} that represent a violation of quite fixed mental images are thus more often recognized as implausible. In contrast, more abstract words that lack a tangible reference object 
seem to open up a greater space of potentially plausible interpretations. This 
possibly invites annotators to cooperate and use their imagination resulting in more plausible ratings for more abstract triples. 

Our findings further suggest that it is the recipient who comes up with an interpretation, thus making sense of the seemingly implausible. Moreover, generating fully implausible events is not trivial, which should be taken into account when using automatically generated implausible triples. 

Lastly, while events based on \textit{s-v-o} triples or comparably simple constructions have been successfully leveraged for exploring selection preference and thematic fit \cite{erk-etal-2010-flexible,zhang-etal-2019-sp,pedinotti-etal-2021-cat}, the addition of context exceeding sentences constructed from \textit{s-v-o} triples could potentially resolve present ambiguity and possibly reduce disagreement. 
We thus encourage future work extending this work by collecting and analyzing plausibility ratings for more complex constructions within broader contexts.


\section{Conclusion}
We presented a novel dataset for physical and abstract plausibility for events in English. Based on naturally occurring sentences extracted from Wikipedia, we infiltrated degrees of abstractness, and automatically generated perturbed pseudo-implausible events. We annotated a filtered and balanced dataset for plausibility using crowd-sourcing and performed extensive cleaning steps to ensure annotation quality. We provided in-depth analyses to explore the relationship between abstractness and plausibility and examined annotator disagreement. 
We hope that the presented dataset is used for both analyzing and modelling the notion of plausibility as well as the exploration of closely related tasks such as selectional preference and thematic fit and relevant downstream tasks including commonsense reasoning, NLI, and coreference resolution. Moreover, we make both raw annotations and a range of aggregations publicly available to foster research on disagreement and enable interpretation from various perspectives.

\section*{Limitations}
In this paper, we present a collection of plausibility ratings for simple sentences in English that are automatically constructed from \textit{s-v-o} triples that are extracted from natural language.  
We are aware that, for example, events such as \textit{eat-skyscraper} might have a plausible interpretation in a given fictional world. When constructing our dataset, we do not explicitly account for triples which might originate from Wikipedia articles with content where other possible worlds are assumed. 


As we conduct a relatively large annotation experiment via AMT crowd-sourcing, we aim to apply post-processing methods minimising the impact of unreliable annotations on our analyses. With more than 500 different final annotators and a very subjective annotation task, we however note the possibility of potentially wrong annotations due to errors, limitations of task instructions, or the interface \cite{pradhan-etal-2012-conll,poesio-etal-2019-crowdsourced,Uma2022ScalingAD}. This is especially true for the implausible portion of the dataset where no comparison with an attested triple label is possible. Approaches of mitigation could be concentrating on triples with high (im)plausibility ratings or use e.g., probabilistic methods to aggregate labels. We thus provide a dataset version with labels aggregated using MACE \cite{hovy-etal-2013-learning}.

As far as the transfer of the suggested approach of dataset construction to languages other than English is concerned, we call attention to the potential need to adapt the event extraction. Further, abstractness ratings might not readily be available in every language. In addition, AMT annotation for languages other than English potentially requires more time and resources, as annotator population is heavily skewed towards speakers of English.

\section*{Ethics Statement}
To generate our dataset of events, we use a portion of the English Wikipedia which has been shown to exhibit a range of biases \cite{olteanu2019,schmahl-etal-2020-wikipedia,falenska-cetinoglu-2021-assessing,sun-peng-2021-men}. While our goal is to enable others to explore plausibility on physical and abstract levels as well as sources of potential disagreement, users of this dataset should acknowledge potential biases and should not use to to make deployment decisions or rule out failures.


In the context of our annotation task, we collected plausibility ratings from crowd-workers using Amazon Mechanical Turk between January, 20 and March 7, 2023. Crowd-workers were compensated 0.02\$ per instance. Although we aimed for strict quality control during data collection, we mostly compensated completed hits also when annotations were finally discarded because they did fail a check instance or, sometimes, did not move the slider. To this end, we engaged in email conversations with crowd-workers in case they reached out to clarify issues. We invested time to answer all requests and made our decision-making transparent to the annotators.

\section*{Acknowledgements}
We are grateful to Prisca Piccirilli, Neele Falk, and the whole SemRel group for helpful suggestions and feedback regarding this work. We would also like to thank the anonymous reviewers for their nuanced comments and constructive suggestions. 
Annerose Eichel received funding by the Hanns Seidel Foundation's Talent Program. 
\bibliography{custom,anthology}
\bibliographystyle{acl_natbib}

\appendix
\section{Dataset Construction} \label{appsec:dataset}

\subsection{Filtering the Brysbaert Norms} \label{appsec:dataset_nltriples}
To assign abstractness scores to all nouns and verbs in a given event triples, we draw on the concreteness ratings for approximately 40,000 English words devised by \citet{brysbaert2014}.
The Brysbaert norms were collected in an out-of-context setting and without providing information about the part-of-speech (POS). POS tags were added in a post-processing step, utilizing the SUBTLEX-US corpus \cite{brysbaert2012}.
To account for this, we follow \citet{SchulteImWalde/Frassinelli:22} and \citet{tater-etal-2022-concreteness} in adding the most frequent POS tag associated with each target word based on the English web corpus ENCOW16AX \cite{schaefer2015}. We then filter for noun and verb target words where the POS tag provided by \cite{brysbaert2014} and the POS tag extracted using the ENCOW16AX correspond to each other. We filter out all words with a frequency below 10K to remove infrequent words. This way, we obtain a collection of 12,880 noun and 2,522 verb targets. 

\subsection{Triple Binning and Distributions} \label{appsec:dataset_distribution} 

\begin{figure*}[htpb]
     \centering
     \begin{subfigure}[b]{0.48\textwidth}
         \centering
         \includegraphics[width=\textwidth]{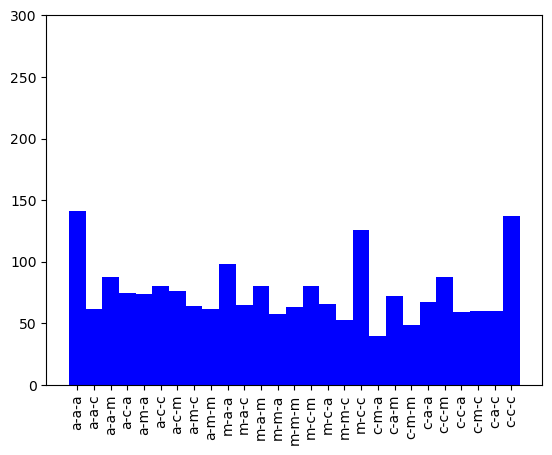}
         \caption{Plausible triples.}
         \label{fig:distribution_plausible_ts}
     \end{subfigure}
     \hfill
     \begin{subfigure}[b]{0.48\textwidth}
         \centering
         \includegraphics[width=\textwidth]{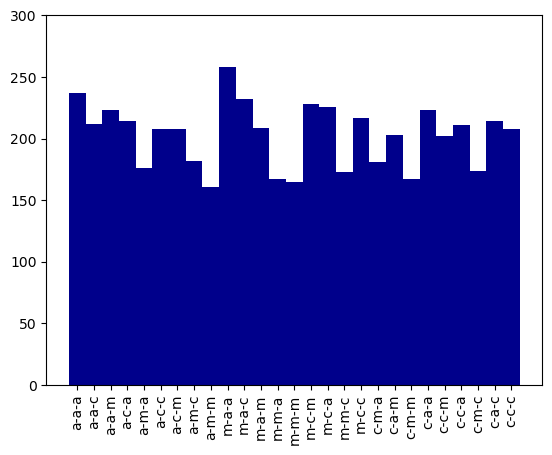}
         \caption{Implausible triples.}
         \label{fig:distribution_implausible_ts}
     \end{subfigure}
     \hfill
        \caption{Distribution of attested plausible (left) and perturbed implausible (right) triples per combination.}
        \label{fig:distributions}
\end{figure*}

The distribution of all naturally occurring triples for each triple combination
$\in \{(a,a,a), ... (c,c,c)\}$ is presented in Fig.~\ref{fig:distributions} (a). While triple numbers accumulate on the extremes highly abstract and highly concrete, the number drops for triples consisting of mid-range constituents. Mixed triple combinations $(a,m,c)$ and $(c,m,a)$ yield minimum numbers of triples as well as triples with highly concrete or abstract subjects and verbs $(a,a,c)$ and $(c,c,a)$.

Similarly, the distribution of all automatically generated pseudo-implausible triples for each triple combination is shown in Fig.~\ref{fig:distributions} (b). Note that a substantially higher number of valid implausible triples is extracted using the binning process with minimum numbers achieved for mostly medium-range abstractness.

\section{Human Annotation}
\label{appsec:human_annotation}

\subsection{HIT Interface} \label{appsec:human_annotation_interface}

\begin{figure*}[!htpb]
    \centering
    \includegraphics[width=\textwidth]{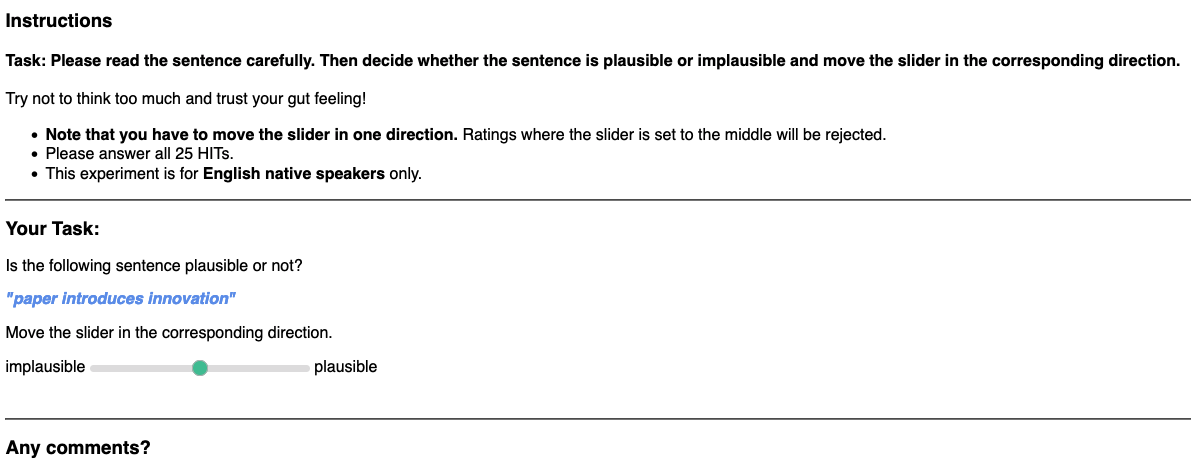}
    \caption{HIT interface including task instruction and requirements for successful answer submission (ratings where the slider is set to the middle can be rejected).}
    \label{fig:hit_example}
\end{figure*}

Fig.~\ref{fig:hit_example} shows a full example of the HIT interface as presented to HIT workers. 

\subsection{Check Instances} \label{appsec:human_annotation_check}

We list check instances in Table~\ref{tab:check_instance}.  In a post-processing step, we exclude three implausible check instances, e.g., \textit{water cuts ball}, which might be interpreted as plausible in the context of high-pressure water systems which might be able to cut a ball (marked in italics). 
We use the check instances mainly after the the annotations process to increase annotation quality by filtering out all submissions where annotators failed a valid check instance. 

\begin{table}[!htpb]
\centering
\small
\begin{tabular}{@{}l|l@{}}
\toprule
\multicolumn{1}{c}{plausible} & \multicolumn{1}{c}{implausible} \\ \midrule
grandmother drinks tea & grandmother drinks stone \\
child eats banana      & child eats dream         \\
baker bakes cake       & \textit{baker bakes air}          \\
kid plays game         & sun beats banana         \\
rabbit eats carrot     & baby eats storm      \\
man builds house       & man breaks air           \\
man opens window       & ant opens window         \\
teenager drinks coke   &  \textit{sun breaks door}         \\
woman drives car       & woman drinks bridge          \\
player throws ball     & \textit{water cuts ball}      \\ \bottomrule
\end{tabular}
\caption{Plausible and implausible check instances. Instances marked in italics are filtered out in a post-processing step due to possible plausible interpretations.}
\label{tab:check_instance}
\end{table}

\subsection{Annotation Post-Processing} \label{appsec:human_annotation_post}

We show an overview of dataset statistics at each post-processing step in Table~\ref{tab:annotation_stats}. Specifically, we present changes in number of ratings, validated annotators, and number of triples with \textgreater{}8 ratings across annotation post-processing. Post-processing methods are applied in the order listed. Results in a given row correspond to dataset statistics having applied a given step.

\paragraph{Soft Jaccard Coefficient} \label{appsec:human_annotation_post}
We estimate Inter-Annotator Agreement (IAA) by calculating the Jaccard Coefficient for all pairwise annotator combinations
%
\[{J(A,B)={\frac {|A\cap B|}{|A\cup B|}}}\]
%
where the intersection of $A$ and $B$ captures all cases where annotators agree that a triple is either plausible (ratings $\in \{4,5\}$) or implausible (ratings $\in \{1,2\}$), and the union of $A$ and $B$ denotes all cases where both annotators provided a rating for the same sentence regardless of their agreement. As we are not enforcing strict rating agreement, we refer to this way of calculating IAA as \textit{soft} Jaccard Coefficient. 




\begin{table*}[!htpb]
\centering
\small
\begin{tabular}{@{}lrr|rrr|r@{}}
\toprule
                                                  & \multicolumn{2}{c}{\# annotators} & \multicolumn{3}{c}{\# ratings} & \# triples\\ \midrule
                                       & plausible & implausible & plausible & implausible & total &   total   \\
Raw (without check instances)          & 883       & 879         & 11,250     & 11,343       & 22,593 & 2,160 \\
Failed checks/default submission       & 878       & 872         & 10,649     & 10,668       & 21,317 & 2,148 \\
\textgreater{}75\% disagreement orig. label & 832             & 838             & 9,849     & 10,046    & 19,895    & 2,081                                  \\
\textless{}10 ratings submitted        & 478       & 479         & 8,373      & 8,502        & 16,875 & 1,840 \\
AMT approv. rate \textless{}80\%, incl. 0\% & 468       & 471         & 8,269      & 8,333        & 16,602 & 1,819 \\
Pairwise Jaccard Index \textless{}0.4       & 452       & 452         & 7,726      & 7,845       & 15,571 & 1,733 \\ \bottomrule
\end{tabular}
\caption{Overview of changes in number of ratings, validated annotators, and number of triples with \textgreater{}8 ratings across annotation post-processing. Post-processing methods are applied in the order listed. Results in a given row correspond to dataset statistics having applied a given step, e.g., filtering out submission from annotators who failed check instances as well as all submissions where annotators submitted a default rating of 3 results in the number of 21,317 valid ratings, including both ratings for plausible and implausible triples. }
\label{tab:annotation_stats}
\vspace{0.15cm}
\end{table*}

\end{document}